\documentclass[runningheads]{llncs}
\usepackage{graphicx}

\usepackage{tikz}
\usepackage{comment}
\usepackage{amsmath,amssymb} 
\usepackage{color}
\usepackage{float}
\usepackage[accsupp]{axessibility}  

\usepackage[width=122mm,left=12mm,paperwidth=146mm,height=193mm,top=12mm,paperheight=217mm]{geometry}
\usepackage{multirow}
\newcommand{\etal}{\textit{et al.}}
\newcommand{\ie}{\textit{i.e.}}
\newcommand{\eg}{\textit{e.g.}}

\usepackage[ruled,vlined]{algorithm2e}
\usepackage{algorithmic}

\usepackage{caption}
\usepackage{xcolor}
\definecolor{commentcolor}{RGB}{110,154,155}   

\usepackage{booktabs}
\usepackage{listings}
\usepackage{xcolor}

\definecolor{codegreen}{rgb}{0,0.6,0}
\definecolor{codegray}{rgb}{0.5,0.5,0.5}
\definecolor{codepurple}{rgb}{0.58,0,0.82}
\definecolor{backcolour}{rgb}{0.95,0.95,0.92}

\lstdefinestyle{mystyle}{
    backgroundcolor=\color{backcolour},
    commentstyle=\color{codegreen},
    keywordstyle=\color{magenta},
    numberstyle=\tiny\color{codegray},
    stringstyle=\color{codepurple},
    basicstyle=\ttfamily\scriptsize,
    breakatwhitespace=false,
    breaklines=true,
    captionpos=b,
    keepspaces=true,
    numbers=left,
    numbersep=5pt,
    showspaces=false,
    showstringspaces=false,
    showtabs=false,
    tabsize=2
}

\begin{document}
\pagestyle{headings}
\mainmatter
\def\ECCVSubNumber{3546}  

\title{BatchFormerV2: Exploring Sample Relationships \\ for Dense Representation Learning}

\titlerunning{BatchFormerV2}
%
\author{Zhi Hou$^1$, Baosheng Yu$^1$, Chaoyue Wang$^2$, Yibing Zhan$^2$, Dacheng Tao$^{2,1}$}
\institute{
$^1$ The University of Sydney, Australia \\
 $^2$ JD Explore Academy, China \\
{\tt\small zhou9878@uni.sydney.edu.au, baosheng.yu@sydney.edu.au, chaoyue.wang@outlook.com, zhanyibing@jd.com, dacheng.tao@gmail.com} \\
}

\maketitle

\begin{abstract}
Attention mechanisms have been very popular in deep neural networks, where the Transformer architecture has achieved great success in not only natural language processing but also visual recognition applications. Recently, a new Transformer module, applying on batch dimension rather than spatial/channel dimension, \ie, BatchFormer~\cite{hou2022batch}, has been introduced to explore sample relationships for overcoming data scarcity challenges. However, it only works with image-level representations for classification. In this paper, we devise a more general batch Transformer module, BatchFormerV2, which further enables exploring sample relationships for dense representation learning. Specifically, when applying the proposed module, it employs a two-stream pipeline during training, \ie, either with or without a BatchFormerV2 module, where the batchformer stream can be removed for testing. Therefore, the proposed method is a plug-and-play module and can be easily integrated into different vision Transformers without any extra inference cost. Without bells and whistles, we show the effectiveness of the proposed method for a variety of popular visual recognition tasks, including image classification and two important dense prediction tasks: object detection and panoptic segmentation. Particularly, BatchFormerV2 consistently improves current DETR-based detection methods (\eg, DETR, Deformable-DETR, Conditional DETR and SMCA) by over {\bf 1.3\%}. Code will be made publicly available.
\end{abstract}

%
%

\section{Introduction}

\begin{figure}[!ht]
\centering
  \includegraphics[width=0.8\linewidth]{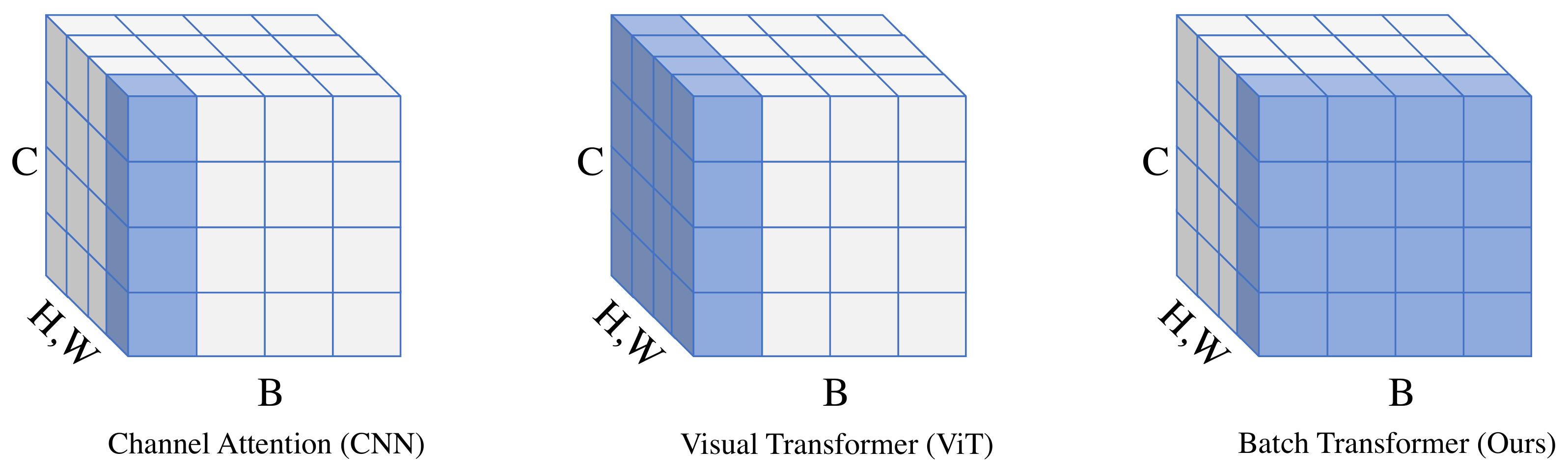}
  \caption{An illustration of the attention mechanisms on channel, spatial, and batch dimensions. Specifically, vision Transformer usually treats image patches as a sequence, and then uses a self-attention layer upon all image patches, \ie, the spatial attention. Different from previous work, we propose to also treat a mini-batch of samples as a sequence, where a typical self-attention layer can thus work on the batch dimension.}
  \label{fig:illu}
\end{figure}

In the past five years, we have witnessed great success of deep neural networks with attention mechanisms, especially the Transformer architecture~\cite{vaswani2017attention}, for natural language processing (NLP). Recently, the Transformer architecture has been successfully extended to visual recognition tasks known as vision Transformer or ViT~\cite{dosovitskiy2020image}, suggesting its great potential as a general deep architecture for learning through different modalities. Therefore, following previous attention mechanisms used in CNNs, such as channel and spatial attentions~\cite{hu2018squeeze,wang2018non,woo2018cbam,carion2020end,gberta_2021_ICML}, recent vision Transformer variants mainly focus on exploring more efficient and effective local/global attention structures~\cite{liu2021swin,yuan2021tokens,chu2021twins,chen2022regionvit}. However, from the perspective of information propagation, current vision Transformer architectures can only enable the spatial information propagation among pixels/patches within the same image, ignoring the possible information propagation among different data samples within each training mini-batch. An intuitive example showing the differences between channel, spatial, and batch attentions is illustrated in Figure~\ref{fig:illu}.


To enable vanilla vision Transformer with the ability of sample relationship modeling, Hou~\etal~present to incorporate a Transformer encoder module between the last hidden layer and the final classifier to explore image-level sample relationships using the attention mechanism, which is referred to as BatchFormer~\cite{hou2022batch} and achieves significant improvements over a wide range of data-scarce image classification tasks. However, BatchFormer mainly focuses on exploring sample relationships at the image-level for classification, leaving sample relationships for dense representations at the pixel/patch level poorly investigated. Inspired by this, we devise a new general batch Transformer module, BatchFormerV2, which further enables the sample relationship learning for dense representations, aiming to facilitate the research on a more powerful and general vision Transformer with not only the spatial attention but also the batch attention, from image-level to pixel/patch-level representation learning.

Exploring sample relationships within each mini-batch using deep neural networks is usually non-trivial, since we do not always have a mini-batch of data for testing, which is also known as the training-and-testing inconsistency. For example, batch normalization requires to track running statistics (\ie, the mean and variance of each mini-batch) during training in a momentum way~\cite{ioffe2015batch}, which is then used as batch statistics for testing. When applying a parametric attention module in the batch dimension, it is even more difficult to handle the above-mentioned inconsistency problem. To this end, Hou~\etal~\cite{hou2022batch} introduce a shared classifier positioned both before and after the batchformer module, aiming to learn a batch-invariant classifier. In this paper, we devise a more general batch Transformer module to further enable the information propagation between samples at different levels, \ie, both pixel/patch and image levels, by learning batch-invariant representations.
Therefore, the proposed batchformer module can be applied to different hidden layers/blocks in typical vision Transformers.
To achieve this, we explore a two-stream pipeline for training, one stream with a batchformer module and the other stream without, where all other layers/blocks are shared by these two streams. By doing this, all these shared layers/blocks are trained to generalize well for the input with or without the batch Transformer module. During testing, we can directly remove the batch Transformer module without sacrificing performance.

In this paper, with the proposed simple yet effective batch Transformer module, we further enable the information propagation along the pixels/patches of different samples within each mini-batch to facilitate dense representation learning. To evaluate the proposed batch Transformer module, named as BatchFormerV2, we perform extensive experiments on several popular visual recognition tasks, including image classification, object detection, and panoptic segmentation, where the experimental results show that the proposed module can be a general representation learning solution via sample relationship learning. The main contributions of this paper are summarized as follows.

\begin{itemize}
  \item We introduce a new plug-and-play module, BatchFormerV2, as a general solution for sample relationship learning at different levels, which further facilitates robust dense representation learning.
  \item We develop a two-stream training strategy, making it possible to improve vision Transformers without any extra inference cost by directly removing the proposed BatchFormerV2 module during testing.
  \item We perform extensive experiments on three popular visual recognition tasks, \textit{i.e.,} image classification, object detection, and panoptic segmentation, demonstrating remarkable generalizability of the proposed method.
\end{itemize}

\section{Related Work}

\subsection{Vision Transformer}
Transformers were first presented by Vaswani \etal \cite{vaswani2017attention} for machine translation based on multi-head self-attention mechanism. As a core part of Transformers, attention mechanism \cite{bahdanau2014neural} aggregates information from the entire input sequence and then update it. In the past several years, Transformer-based architectures have dominated in natural language processing (NLP). For example, large-scale Transformer-based models, \eg, BERT~\cite{devlin2018bert}, show superior performance among massive down-stream NLP tasks. Besides, self-attention also demonstrates the powerful modeling ability of non-grid data~\cite{velickovic2017graph}, and improves graph representation learning.

Recently, Transformer models present new paradigms for computer vision tasks, including classification~\cite{dosovitskiy2020image,liu2021swin}, detection~\cite{carion2020end,zhu2020deformable,meng2021-CondDETR,gao2021fast}, segmentation~\cite{carion2020end,cheng2021per,zheng2020rethinking,strudel2021,wang2021pyramid,xie2021segformer}, and representation learning~\cite{chen2021empirical,bao2021beit,he2021masked}. Specifically, Dosovitskiy \etal~\cite{dosovitskiy2020image} introduced a pure Transformer model, termed as ViT, to apply a sequence of image patches and achieve comparable performance on image classification tasks. Recently, Visual Transformer~\cite{dosovitskiy2020image,liu2021swin} has gradually become a new backbone for visual tasks, and massive large models based on Transformers have emerged in computer vision, including CLIP~\cite{radford2021learning}, MoCo~\cite{he2020momentum,chen2021empirical}, DINO~\cite{caron2021emerging}, DALL-E~\cite{ramesh2021zero}, BEiT~\cite{bao2021beit}, and MAE~\cite{he2021masked}. Except for the backbone, Transformer-based models, \eg, DETR~\cite{carion2020end}, have also reformed the pipeline of detection and segmentation. DETR~\cite{carion2020end}, constructed upon the encoder-decoder Transformer architecture, demonstrates a clear set-based method for detection and greatly simplifies the traditional pipeline which includes many hand-designed components. Recently, Zhu \etal ~\cite{zhu2020deformable} present Deformable DETR, which largely accelerates the convergence and improves the performance.

Though the great success of Transformer in computer vision, current approaches merely investigate the spatial self-attention, while ignoring the pixel/patch relationships among samples. Recently, Hou~\etal~\cite{hou2022batch} present to explore the sample relationships in the image level for data scarcity challenges, which however is impractical to dense prediction tasks. In this paper, we introduce a novel Batch Transformer, BatchFormerV2, to enable information propagation among samples in the pixel/patch level, and largely facilitate dense representation learning for dense prediction tasks, \eg, Object Detection and Panoptic Segmentation.


\subsection{Sample Relationship Learning}
Mini-Batch stochastic gradient descent optimization is one of core paradigm of training deep neural networks~\cite{goodfellow2016deep,krizhevsky2012imagenet,lecun2015deep}. In order to accelerate the network training, Ioffe \& Szegedy~\cite{ioffe2015batch} present Batch Normalization to reduce internal covariate shift via normalizing intermediate representations. Batch Normalization has inspired a large number of cross-batch techniques~\cite{wu2018group,chang2019domain}, and has become one of cornerstones for modern deep networks. Recently, Hou~\etal~\cite{hou2022batch} introduce a BatchFormer module in the penultimate layer with a shared classifier for data scarcity tasks. Inspired by them, BatchFormerV2 proposed in this paper is pluggable into different layers in Visual Transformer and DETR, which first enables applying batch Transformer module for dense prediction tasks. Here, we compare the proposed BatchFormerV2 with other cross-batch techniques. First, Batch Normalization actually enables the sample information propagation in a linear way, and stables the network training. Differently, we think the proposed BatchFormerV2 propagates information among samples in the mini-batch in a non-linear way via multi-head self-attention. Meanwhile, transductive inference~\cite{liu2018learning,satorras2018few} is also popular in few-shot approaches and recent approaches~\cite{kossen2021self,mondal2021mini} introduce to reason the relationships between samples, while those techniques require batch inference and thus related applications are limited. In addition, Mixup~\cite{zhang2017mixup} is a popular augmentation technique, which draws virtual samples from the vicinity distribution of the training examples. Similar to Mixup~\cite{zhang2017mixup}, our BatchFormerV2 also changes the feature space according to the mini-batch training samples. Differently, BatchFormerV2 is able to insert different layers in Transformer modules, and learns the transform implicitly. Therefore, BatchFormerV2 can be easily applied to dense prediction tasks.

\section{Method}

In this section, we first revisit the attention mechanisms in vision Transformer. We then introduce the proposed BatchFormerV2. Lastly, we describe the two-stream training pipeline for the optimization of vision Transformers with the proposed method.

\subsection{A Revisit of Vision Transformer}
The great success of Transformer architecture in NLP has recently spread to almost every region of computer vision known as vision Transformers~\cite{dosovitskiy2020image,liu2021swin,carion2020end}. The Transformer architecture not only achieves superior performance in different vision tasks~\cite{liu2021swin} (\eg, image classification, object detection, and semantic segmentation), but also bring novel paradigms for some fundamental tasks (\eg,  DETR~\cite{carion2020end} for object detection and MAE~\cite{he2021masked} for self-supervised learning). Among typical vision Transformers, the overall model usually consists of a stack of multiple Transformer encoder blocks~\cite{vaswani2017attention}, where each Transformer encoder block contains a multi-head self-attention layer (MHSA) followed by a feed-forward network (FFN). Specifically, the self-attention mechanism used in vision Transformer can be described as follows. Given $\mathbf{Q},~\mathbf{K},~\mathbf{V}\in R^{N\times C}$ as the query, key, and value, respectively, where $N$ is the number of image patches (or tokens) and $C$ is the embedding dimension. We then have the output $\mathbf{Z}$ for the self-attention module:
\begin{equation*}
    \mathbf{Z} = \text{softmax}(\frac{\mathbf{Q}\mathbf{K}^\top}{\sqrt{C}})\mathbf{V},
\end{equation*}
where $\mathbf{Q}$, $\mathbf{K}$ and $\mathbf{V}$ are learned from the same input. Specifically, multi-head self-attention module applies attention by spliting the input into multiple representation subspaces and then concatenates the representations from different heads. From a perspective of information propagation, the Transformer architecture aggregates the feature of the tokens via spatial attention. Different from typical attention layers used in vision Transformer, BatchFormer performs a self-attention on the batch dimension, \ie, it aggregates the feature of the tokens from different samples within each mini-batch in an end-to-end learning way.

\subsection{BatchFormerV2}

\begin{figure}[!ht]
    \centering
    \includegraphics[width=.85\textwidth]{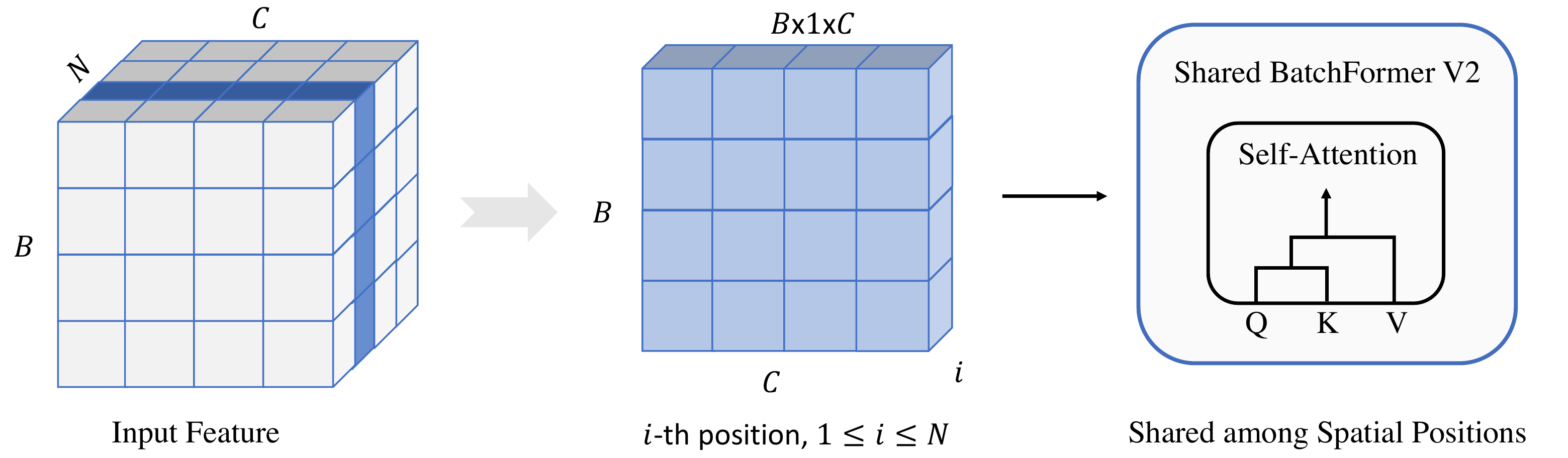}
    \caption{Illustration of BatchFormerV2. For the $i$-th spatial position in the feature map, we apply a Transformer block to the feature map where the length of each sequence is the batch size.}
    \label{fig:pbt}
\end{figure}

To generalize the batch attention mechanism into pixel/patch level feature maps for dense representation learning, we devise BatchFormerV2 as follows. Given $\mathbf{Q},~\mathbf{K},~\mathbf{V} \in R^{B\times N\times C}$, we then have
\begin{equation}
    \mathbf{Z}_i = \text{softmax}(\frac{\mathbf{Q}_{i}\mathbf{K}_i^{\top}}{\sqrt{C}})\mathbf{V}_i,~~~\mathbf{Z} = \text{concat} (\mathbf{Z}_1,\dots,\mathbf{Z}_N),
\end{equation}
where $\mathbf{Q}_i,~\mathbf{K}_i,~\mathbf{V}_i \in R^{B\times C}$ and $\mathbf{Z}\in R^{B\times N\times C}$.
As illustrated in Figure~\ref{fig:pbt}, given the input for a specific layer/block with the spatial dimensions $H,~W$, \ie, the number of image patches is $N=H\times W$. During training, at each spatial position $i=1,\dots, N$, we treat the batch of patch features in current position as a sequence, \ie, we have $N$ sequences each with the length of $B$. All above-mentioned sequences are then feed into a shared Transformer block.

Two reasons for using a shared Transformer block are as follows: 1) it will increase the computation and memory consumption considerably if we use different blocks at different spatial positions; 2) it will be difficult to dense prediction with different sizes of input images, which is in line with the motivation of fully convolutional networks (FCNs) for dense prediction~\cite{long2015fully} as well as the convolution operations~\cite{lecun1995convolutional} and channel-wise attentions~\cite{hu2018squeeze}.
Therefore, we share the Transformer block among the spatial dimensions in BatchFormerV2. In addition, by doing this, the proposed BatchFormerV2 can be implemented by simply transposing the spatial and batch dimensions before the standard multi-head self-attention layers. Noticeably, as illustrated in Figure~\ref{fig:alg}, BatchFormerV2 can be easily implemented with a few lines of code using popular deep learning packages such as PyTorch~\cite{paszke2019pytorch}.


\begin{figure}[!ht]
\centering
\begin{lstlisting}[language=Python]
def batch_former_v2(x, encoder, is_training, is_first_layer):
    # x: input features with the shape (B, N, C).
    # encoder: TransformerEncoderLayer(C, nhead, C, 0.5, batch_first=False)
    if not is_training:
        return x
    orig_x = x
    if not is_fist_layer:
        orig_x, x = torch.split(x, len(x)//2)
    x = encoder(x)
    x = torch.cat([orig_x, x], dim=0)
    return x
\end{lstlisting}
\caption{Python code of BatchFormerV2 based on PyTorch.}
\label{fig:alg}
\end{figure}

\subsection{Two-Stream Training}

\begin{figure}[!ht]
    \centering
    \includegraphics[width=.95\textwidth]{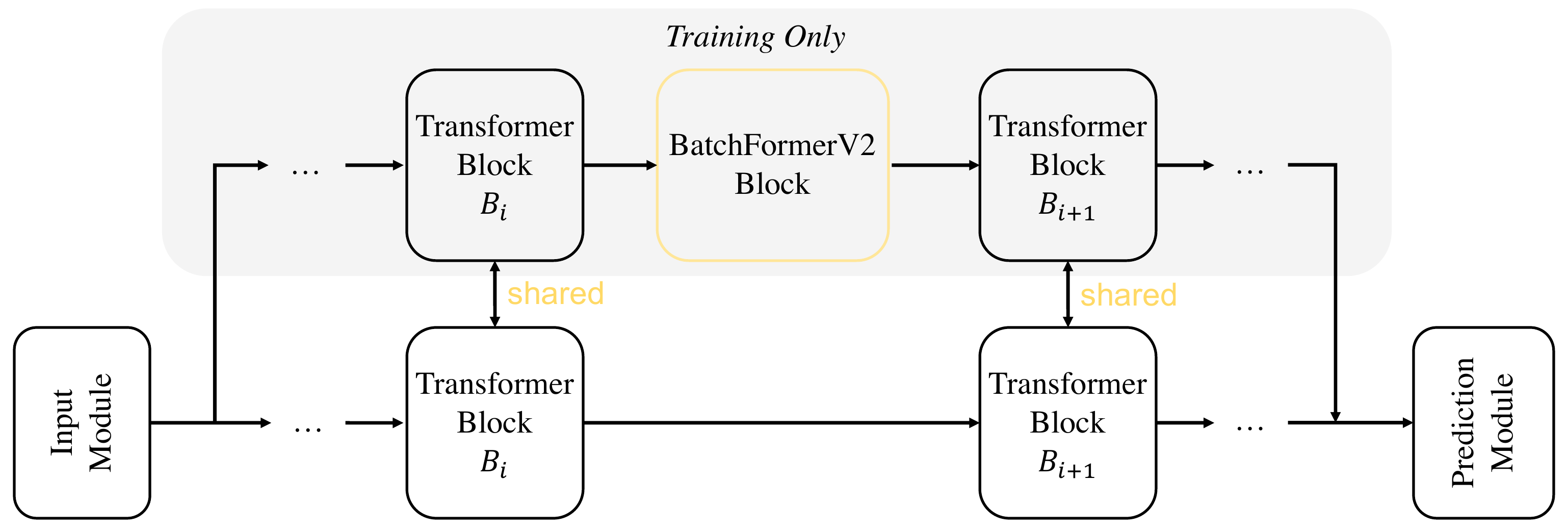}
    \caption{The two-stream training pipeline for the proposed BatchFormerV2. For example, here the input for DETR~\cite{carion2020end} indicates the feature map from the backbone network, while the input for ViT~\cite{dosovitskiy2020image} is the feature map after the patch embedding layer. We copy the feature and input into BatchFromerV2 stream. The outputs of the two streams are inputted into shared prediction module. Particularly, the Transformer Blocks and prediction module, \eg, Transformer Decoder in DETR~\cite{carion2020end} and classification head in ViT~\cite{dosovitskiy2020image}, are shared between the two streams. Noticeably, during inference, the BatchFormer stream is removed and thus there are no additional computation.}
    \label{fig:arch}
\end{figure}

One of the most significant challenge for applying batch attention is the inconsistency between training and testing~\cite{kossen2021self,hou2022batch}. Specifically, in BatchFormer~\cite{hou2022batch}, Hou~\etal~ address the inconsistency problem by introducing a shared classifier, which enables to remove BatchFormer module during inference. Inspired by this, we generalize this solution to dense representation learning by utilizing the similar idea, \ie, batch-invariant learning. Therefore, we introduce a new two-stream training strategy for using BatchFormerV2 as follows.

When applying the proposed BatchFormerV2 module to a specific block of vision Transformers, we create a new siamese stream followed by a BatchFormerV2 module, leaving the original stream unchanged. That is, both two streams share the same Transformer block. By doing this, during training, all shared blocks are trained on a mixture of the distributions with or without a BatchFormerV2 module. Therefore, during testing, the original stream can work well for both with and without a mini-batch of testing data available. To avoid introducing any extra inference load, we thus remove the BatchFormerV2 module for testing. In addition, from the perspective of regularization, the proposed BatchFormerV2 also serves as a strong regularization during training, which has turned out to be very useful in vision Transformers.
Lastly, with the proposed two-stream training strategy, a BatchFormerV2 module can be easily integrated into existing Transformer architectures for different visual applications, such as ViT~\cite{dosovitskiy2020image} for image classification and DETR~\cite{carion2020end}/Deformable-DETR~\cite{zhu2020deformable} for object detection, in a plug-and-play manner.


\section{Experiments}

In this section, we evaluate the proposed BatchFormerV2 module for dense representation learning on popular dense prediction tasks, including object detection and panoptic segmentation. We also perform comprehensive ablation studies on common object detection. Lastly, as a more general solution than vanilla BatchFormer, we also perform experiments on image classification, demonstrating the generalizability of BatchFormerV2. Please refer to Appendix for the effectiveness of BatchFormerV2 on Long-Tailed Recognition, 3D Hand Mesh Reconstruction, and Self-Supervised Learning.

\subsection{Object Detection}

{\bf Implementation Details}.
We use the popular DETR/Deformable-DETR as our baseline methods, and apply the proposed BatchFormerV2 for object detection. We perform all experiments on the most popular common object detection benchmark dataset, COCO 2017~\cite{lin2014microsoft}, which contains 118k training images and 5k validation images. During training, the backbone network is initialized from the weights pretrained on ImageNet-1K~\cite{deng2009imagenet}.
We run experiments on eight Nvidia V100 GPUs (16GB) for DETR, and eight Nvidia A100 GPUs (40GB) for Deformable DETR. If not otherwise stated, the batch-size for DETR, Conditional DETR and SMCA is 16 and the default batch size for Deformable DETR is 24. If not stated, we insert the BatchFormer module in the first Transformer encoder layer in experiments. The number of heads in BatchFormerv2 is 4. For fair comparisons, all other hyperparameters follow the default configurations described in DETR~\cite{carion2020end},Conditional DETR~\cite{meng2021-CondDETR}, SMCA~\cite{gao2021fast} and Deformable DETR~\cite{zhu2020deformable}. An ablation study on the batch size can also be found in Section~\ref{sec:ab_obj}.

\setlength{\tabcolsep}{5.5pt}
\begin{table}
\begin{center}
\begin{tabular}{@{}lccccccc@{}}
\toprule
Method & Backbone & $AP$ & $AP_{50}$ & $AP_{75}$ & $AP_S$ & $AP_M$ & $AP_L$ \\
\hline
DETR~\cite{carion2020end}& ResNet-50  & 34.8 & 55.6 & 35.8 & 14.0 & 37.2 & 54.6 \\
+ BatchFormerV2 & ResNet-50 & {\bf 36.9} & {\bf 57.9}  & {\bf 38.5} &  {\bf 15.6} & {\bf 40.0} & {\bf 55.9 }\\
\hline
Conditional DETR~\cite{meng2021-CondDETR} & ResNet-50 & 40.9 & 61.8 & 43.3 & 20.8 & 44.6 & 59.2 \\
+BatchFormerV2 & ResNet-50 & {\bf 42.3} & {\bf 63.2} & {\bf 45.1} & {\bf 21.9} & {\bf 46.0} & {\bf 60.7} \\
\hline
SMCA (single scale)~\cite{gao2021fast} & ResNet-50 & 41.0 & - & - & 21.9 & 44.3 & 59.1 \\
+BatchFormerV2 & ResNet-50 & {\bf 42.3} & {\bf 63.5} & {\bf 45.4} & {\bf 22.5} & {\bf 45.7} & {\bf 60.1} \\
\hline
Deformable DETR~\cite{zhu2020deformable}& ResNet-50 & 43.8 & 62.6 & 47.7 & 26.4 & 47.1 & 58.0 \\
+ BatchFormerV2 & ResNet-50  & {\bf 45.5} & {\bf 64.3} & {\bf 49.8} & {\bf 28.3} & {\bf 48.6} & {\bf 59.4} \\
\hline
Deformable DETR$^*$~\cite{zhu2020deformable}& ResNet-50 & 45.4 & 64.7 & 49.0 & 26.8 & 48.3 & 61.7\\
+ BatchFormerV2 & ResNet-50  & {\bf 46.7} & {\bf 65.6} & {\bf 50.5} & {\bf 28.8} & {\bf 49.7} & {\bf 61.8} \\
\hline
Deformable DETR~\cite{zhu2020deformable}& ResNet-101  & 44.5 & 63.7 & 48.3 & 25.8 & 48.6 & 59.6 \\
+ BatchFormerV2 & ResNet-101  & {\bf 46.0} & {\bf 65.2} & {\bf 50.5} & {\bf 28.4} & {\bf 49.8} & {\bf 60.7} \\
\bottomrule
\end{tabular}
\end{center}
    \caption{Illustration of BatchFormerV2 on common object detection based on Deformable DETR~\cite{zhu2020deformable} and DETR~\cite{carion2020end}. Following Deformable DETR~\cite{zhu2020deformable}, we train all models with 50 epochs using the released code. Deformable DETR$^*$ indicates Deformable DETR with iterative bounding box refinement.}
\label{tab:obj_det}
\end{table}

\noindent {\bf Results}. As shown in Table~\ref{tab:obj_det}, BatchFormerV2 significantly improves the corresponding baseline methods. For example, without bells and whistles, BatchFormerV2 improves DETR by 2.1\% and Deformable DETR by 1.7\% when using a ResNet-50 backbone. We observe consistent improvement on Conditional DETR and SMCA. Moreover, we find that BatchFormerV2 mainly improves the object detection performance on small and medium objects. For example, BatchFormerV2 increases Deformable DETR in APS by 1.9\% and APM by 1.5\%, respectively. For DETR, BatchFormerV2 increases $AP_S$ and $AP_M$ by 1.6\% and 2.8\%, respectively. We think via building Transformer along the pixel of the feature map in the batch dimension, BatchFormerV2 utilizes features from other images to facilitates object detection in current image. For small objects which is usually challenging to detect, BatchFormerV2 is able to incorporate objects from other images to detect (refer to the Visualization in Section~\ref{sec:ab_obj}). Therefore, BatchFormerV2 significantly improves corresponding baselines.

\subsection{\bf Ablation Studies on Object Detection}
\label{sec:ab_obj}

To better understand the proposed method for dense representation learning, we perform ablation studies on some key factors that may have influences on the BatchFormerV2 performance. Please also refer to supplementary materials for more results.


\begin{table}
\begin{center}
\setlength{\tabcolsep}{6pt}
\begin{tabular}{@{}cccccccc@{}}
\toprule
Batch Size & Epochs & $AP$ & $AP_{50}$ & $AP_{75}$ & $AP_S$ & $AP_M$ & $AP_L$ \\
\hline
16 & 50 & 44.7 & 63.5 & 48.9 & 27.3 & 48.1 & 59.1 \\
24 & 50 & {\bf 45.1} & {\bf 64.1} & {\bf 49.3} & {\bf 28.5} & {\bf 48.4} & 59.4 \\
32 & 50 & 44.9 & 63.8 & 48.8 & 27.7 & 48.3 & {\bf 60.0} \\
\bottomrule
\end{tabular}
\end{center}
\caption{Ablation study on the training mini-batch size. We use Deformable DETR as our baseline and insert BatchFormerV2 module in the last Transformer layer.}
\label{tab:ab_obj_bs}
\end{table}

\noindent\textbf{Mini-Batch Size}. Considering that BatchFormerV2 aims to learn sample relationships among each mini-batch during training, we evaluate the influence of different mini-batch size on BatchFormerV2 as follows. As shown in  Table~\ref{tab:ab_obj_bs}, 1) when increasing the batch-size from 16 to 24, the performance can be further improved with a small margin; 2) when further increasing the batch-size to 32, the performance is comparable, \ie, no additional improvements. Here, we maintain other hyper-parameters when increasing the batch size. We consider that it may require to tune other hyperparameters after increasing the batch size to achieve further improvements.

\begin{table}
\setlength\tabcolsep{4.pt}
\begin{center}
\begin{tabular}{@{}lccccccccccc @{}}
\toprule
& L1-2 & L1-3 & L3-6 & L4-6 & L5-6 & L1 & L2 & L3 & L4 & L5 & L6 \\
\hline
Accuracy(AP) & 45.2 & 44.9  & 45.4 & 44.9 & 45.0  & {\bf 45.5} & 45.3 & 45.2 & 45.2 & 45.2 & 45.1 \\
\bottomrule
\end{tabular}
\end{center}
\caption{Ablation study on the insert position. Specifically, ``L1-3'' indicates that we insert BatchFormerV2 modules from the first layer to the third layer. ``L1" indicates that we only insert a BatchFormerV2 model in the first layer. Since the GPU memory limitation, we did not insert BatchFormerV2 in all layers. Instead, we selected some combinations that include all layers.}
\label{tab:ab_obj_layers}
\end{table}

\noindent{\bf Insert Position}. We show the influence of different insert positions for BatchFormerV2. Specifically, we use Deformable DETR as the baseline, which contains six Transformer layers. As shown in Table~\ref{tab:ab_obj_layers}, we find that: 1) the insert positions do have an important effect on the performance; and 2) the number of BatchFormer modules does not have significant influence on the final performance, \ie, more BatchFormerV2 modules cannot further improve the object detection performance; and 3) inserting BatchFormerV2 modules in early layers seems to be more effective for dense prediction tasks.



\setlength{\tabcolsep}{4pt}
\begin{table}
\begin{center}
\begin{tabular}{@{}lccccccc@{}}
\toprule
Method & Epochs & $AP$ & $AP_{50}$ & $AP_{75}$ & $AP_S$ & $AP_M$ & $AP_L$ \\
\hline
Deformable DETR & 50 & 43.8 & 62.6 & 47.7 & 26.4 & 47.1 & 58.0 \\
+ BatchFormerV2 (shared) & 50 & 44.9 & 63.6&  49.1 & 27.7 & 47.9 & 59.6 \\
+ BatchFormerV2 (non-shared) & 50 & {\bf 45.4} & {\bf 64.3} & {\bf 49.5} & {\bf 28.6} & {\bf 48.5} & {\bf 59.9} \\
\bottomrule
\end{tabular}
\end{center}
\caption{Ablation study on shared BatchFormerV2 modules. Specifically, we show the influence when sharing the same BatchFormerV2 module among different layers. Here, we insert BatchFormerV2 modules from the third layer to the sixth layer.}
\label{tab:ab_obj_shared}
\end{table}

\noindent\textbf{Shared Modules}.
Furthermore, we test whether sharing or not sharing the BatchFormerV2 module among different layers would benefit the dense preidiction tasks. As shown in~Table~\ref{tab:ab_obj_shared}, we find that not-shared BatchFormerV2 could bring 0.5\% improvement compared to the shared scheme. It may suggest that the dense sample relationships are varying among different layers/levels, which also explains that, for different vision recognition tasks, the proposed BatchFormerV2 may add into different layers.

\begin{figure}
    \centering
    \includegraphics[width=.99\textwidth]{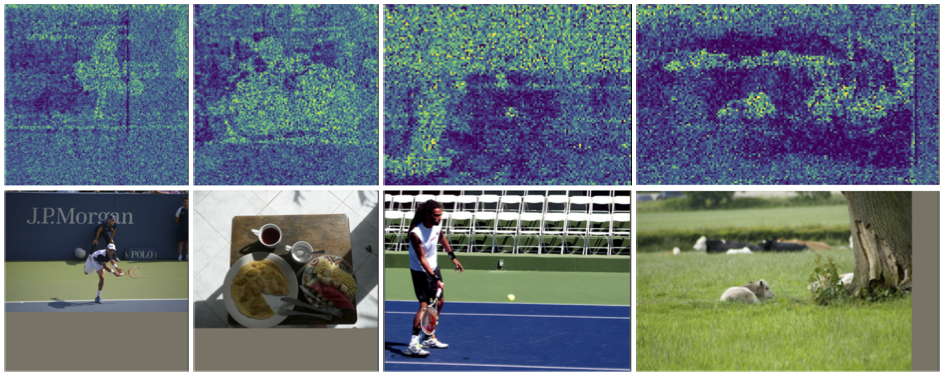}
    \caption{Visualization of self-attention maps. Here, we insert a BatchFormerV2 module into the first Transformer layer in Deformable DETR.}
    \label{fig:bt_atten}
\end{figure}


\noindent\textbf{Visualization}.
BatchFormerV2 are inserted into Transformer Encoder layers along each spatial position in the batch dimension, it enables the information propagation among samples in a mini-batch via the attention mechanism. Here, we visualize the attention of each slot on other slots along with the same position among the batch dimension. Particularly, we choose the largest scale feature map in Deformable DETR in the first Transformer Encoder layer to visualize.
Figure~\ref{fig:bt_atten} shows BatchFormerV2 mainly focuses on the objects (\eg, person, chairs), while paying less attention on the background (\eg, play ground, grass). This actually demonstrates BatchFormerV2 improves the object localization via the self-attention among different samples. We think this is verified to the result of Panoptic Segmentation (Section \ref{sec:seg}) in which the improvement of BatchFormerV2 on the segmentation of things is significantly better than that on the stuff segmentation. In addition, we observe self-attention highlights all the regions of the objects with clear boundaries, which explains the significant improvement of BatchFormerV2 on panoptic segmentation. Last but not least, in Figure~\ref{fig:bt_atten} the mini-batch samples in the test set are randomly constructed, which illustrates objects from different categories can be also mutually beneficial for the localization. Please refer to Appendix for more visualized comparisons.

\subsection{Panoptic Segmentation}
\label{sec:seg}

\textbf{Implementation Details}. We evaluate the proposed BatchFormerV2 for panoptic segmentation, which is a combination of instance and semantic segmentation tasks, on MS-COCO dataset. Specifically, we use the panoptic annotation provided by~\cite{kirillov2019panoptic}, which contains 53 \textit{stuff} categories in addition to 80 \textit{things} categories from the original MS-COCO dataset. We use DETR~\cite{carion2020end} as our baseline for panoptic segmentation, \ie, we utilize a mask head to generate panoptic segmentation results by treating both \textit{stuff} and \textit{things} classes in a unified way~\cite{kirillov2019panoptic}. Following~\cite{carion2020end}, we first train the model with BatchFormerV2 modules for object detection to predict bounding boxes around \textit{stuff} and \textit{things} classes 300 epochs. After that, we finetune the new mask head for extra 25 epochs.

\setlength{\tabcolsep}{4pt}
\begin{table}
\small
\setlength\tabcolsep{3pt}
\begin{center}
\begin{tabular}{@{}l|ccc|ccc|ccc|c@{}}
\toprule
Method &  $PQ$ & $SQ$ & $RQ$ & $PQ^{th}$ & $SQ^{th}$ & $RQ^{th}$ & $PQ^{st}$ & $SQ^{st}$ & $RQ^{st}$ & $AP$ \\
\hline
DETR~\cite{carion2020end} & 43.4 & 79.3&  53.8 &48.2 &79.8 &59.5& 36.3 &78.5 & 45.3 & 31.1 \\
+BatchFormerV2 & {\bf 45.1} & {\bf 80.3} & {\bf 55.3} & {\bf 50.5} & {\bf 81.1} & {\bf 61.5} & {\bf 37.1} & {\bf 79.1} & {\bf 46.0} & {\bf 33.4} \\
\bottomrule
\end{tabular}
\end{center}
\caption{Panoptic segmentation with DETR~\cite{carion2020end} on the COCO val dataset. $PQ^{th}$ and $PQ^{st}$ indicate the results on \textit{things} and \textit{stuff} classes, respectively.}
\label{tab:def_detr_seg}
\end{table}


\noindent\textbf{Results}. We report the panoptic quality ($PQ$) and the breakdown performances on things ($PQ^{th}$) and stuff ($PQ^{st}$)  in Table~\ref{tab:def_detr_seg}. Specifically, we observe that BatchFormerV2 significantly improves $AP$ by 2.3\% and $PQ$ by 1.7\%. We also notice the improvement on $PQ^{th}$ is much larger than $PQ^{st}$. That is, BatchFormerV2 improves $PQ^{th}$ by 2.3\%, while the improvement on $PQ^{st}$ is only 0.8\%. This result is consistent with the results of object detection: by enabling the information propagation, BatchFormerV2 mainly facilitates object detection and instance segmentation.
Furthermore, following~\cite{carion2020end},  we actually freeze the bounding box branch and Transformer layers (include BatchFormerV2) when finetuning the mask head, we find that the performance of panoptic segmentation is also significantly improved. A possible explanation is that BatchFormerV2 improves the optimization of the backbone and the Transformer encoder for better object detail modeling for bounding box detection and subsequently facilitates the segmentation performance when finetuning the mask head.

\begin{figure}[H]
    \centering
    \includegraphics[width=\linewidth]{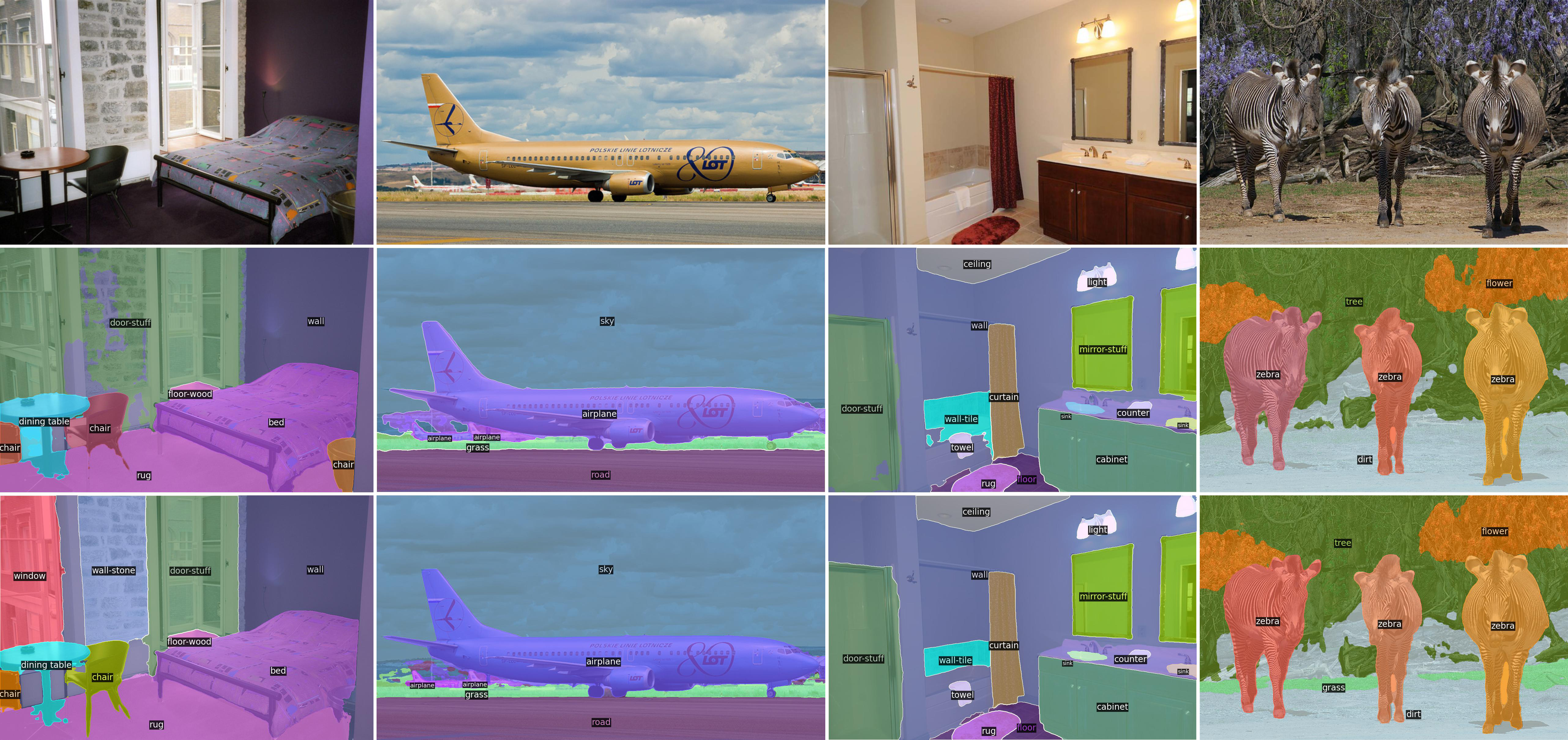}
    \caption{A visual comparison between DETR with and without using the proposed BatchFormerV2. Specifically, the first row is the original image, the second row is the panoptic segmentation result without BatchFormerV2, and the last row indicates the panoptic segmentation result with BatchFormerV2.}
    \label{fig:panoptic_result}
\end{figure}

\noindent\textbf{Visualization}. We find BatchFormerV2 improves the details of segmentation and improves the segmentation on small objects. For example, we find with BatchFormer, the segmentation boundary of door is significantly improved, while the baseline model (\ie, DETR~\cite{carion2020end}) segments the door as wall by mistaking in the left area of Figure~\ref{fig:panoptic_result}. Meanwhile, the legs of the desk are also more clear. In the second figure, we can find the segmentation of airplane achieves better details (\eg, the tail and front wheels) with BatchFormerV2. In the last subfigure, the baseline model ignores the segmentation of grass (\ie, small stuffs), while model with BatchFormerV2 can correctly segment the grass.

\subsection{Image Classification}


\noindent {\bf Implementation Details}. We evaluate BatchFormerV2 for image classification using the most popular visual Transformer (ViT) using the same training strategy with DeiT~\cite{touvron2021deit}. We perform image classification on two popular benchmarks, \ie, CIFAR-100 and ImageNet. Specifically, we train all models for 300 epochs on ImageNet with an initial learning rate 1e-3 and the batch size 1024. For CIFAR-100, following~\cite{chen2021empirical}, we train all models with the initial learning rate 6e-4 and the batch size 1024. All experiments are conducted on a cluster with eight Nvidia A100 GPUs (40GB). When applying the proposed BatchFormerV2, we use the same  number of heads with the corresponding baseline model. For ImageNet dataset, we insert BatchFormerV2 in the eighth layer. Emperically, we observe frequent crashes during training if we insert BatchFormerV2 module in very early layers. For CIFAR-100, we insert BatchFormerV2 module for all layers. More details and analysis are provided in supplementary materials.

\begin{table}
\begin{center}
\setlength{\tabcolsep}{6pt}
\begin{tabular}{@{}lcccc@{}}
\toprule
Model & Params & Input & Top-1 & Top-5 \\
\hline
DeiT-Ti~\cite{touvron2021deit} & 5M & 224$^2$ & 72.2 & 91.1\\
+ BatchFormerV2& 5M & 224$^2$ & {\bf 72.7} & {\bf 91.5}\\
\hline
DeiT-S~\cite{touvron2021deit} & 22M & 224$^2$ & 79.8 & 95.0 \\
+ BatchFormerV2& 22M & 224$^2$ & {\bf 80.4} & {\bf 95.2} \\
\hline
DeiT-B~\cite{touvron2021deit} & 86M & 224$^2$ & 81.7 & 95.5 \\
+ BatchFormerV2 & 86M & 224$^2$ & {\bf 82.2} & {\bf 95.8}\\
\bottomrule
\end{tabular}
\end{center}
\caption{Image classification on ImageNet. We follow the same experimental setups described in DeiT~\cite{touvron2021deit}.}
\label{tab:imgnet}
\end{table}

\noindent\textbf{Results on ImageNet}. Table~\ref{tab:imgnet} demonstrates BatchFormerV2 consistently improves the performance among different ViT models. We observe BatchFormerV2 achieves similar improvement, \ie, around 0.5\%, compared to the baseline. Compared to the improvement on object detection and panoptic segmentation, the improvement on image classification is relatively small. It might be because dense prediction requires to localize the objects in the images, \ie, there are multiple targets in the image, while image classification treats the whole image as the target and requires to recognize the image. Furthermore, we think the strong data augmentation in classification might be also a limitation for BatchFormerV2 on ImageNet. We also analyze the effect of cutmix~\cite{yun2019cutmix} and mixup~\cite{zhang2018mixup} on BatchFormerV2 in supplementary materials. \\

\begin{table}
\begin{center}
\setlength{\tabcolsep}{6pt}
\begin{tabular}{@{}lcccc@{}}
\toprule
Model & Params & Input & Epochs=100 &  Epochs=300 \\
\hline
DeiT-Ti~\cite{touvron2021deit} & 5M & 224$^2$ & 49.2 &  69.2 \\
+ BatchFormerV2& 5M & 224$^2$ & {\bf 58.7} & {\bf 73.4} \\
\hline
DeiT-S~\cite{touvron2021deit} & 22M & 224$^2$ & 57.5 &  72.5 \\
+ BatchFormerV2& 22M & 224$^2$ & {\bf 68.5} &  {\bf 75.2} \\
\hline
DeiT-B~\cite{touvron2021deit} & 86M & 224$^2$ &  52.2 & 71.8 \\
+ BatchFormerV2 & 86M & 224$^2$ &  {\bf 66.6} &  {\bf 74.8}\\
\bottomrule
\end{tabular}
\end{center}
\caption{Image classification on CIFAR-100. Following the experimental setups in~\cite{chen2021empirical}, we train all models from scratch and report the top-1 accuracy (\%). Specifically, DeiT-Ti, DeiT-S, and DeiT-B have the same architectures with ViT-Ti, ViT-S, and ViT-B, respectively.}
\label{tab:cifar100}
\end{table}

\noindent\textbf{Results on CIFAR-100}. Current vision Transformer architectures (\eg, ViT~\cite{dosovitskiy2020image}) usually require a large amount of training data or strong regularization to avoid severe overfitting problems. Therefore, it is still chanllenging for vision Transformer to train from scratch on a small dataset. In this paper, we also find that the proposed BatchFormerV2 module can significantly improves the performance of vision Transformer on small datasets. As illustrated in Table~\ref{tab:cifar100}, BatchFormerV2 significantly improves the performance of DeiT-B from 52.2\% to 66.6 \% by {\bf 14.4\%}, DeiT-S from 57.5\% to 68.5\% by {\bf 11\%}, DeiT-Ti from 49.2\% to 58.7\% by {\bf 9.5\%}. When we train all models with more epochs, \ie, 300 epochs,  the improvement is also considerable. The possible reason is that BatchFormerV2 enables the information propagation among patches in different images, which benefits the optimization and generalization when learning on small datasets. Particularly, we find that DeiT-B does not achieve better performance compared to DeiT-S. This is possibly because DeiT-B is a too large model for a very small dataset, \eg, CIFAR-100.


\section{Limitations and Future Work}
Though the proposed BatchFormerV2 module is very effective and efficient for common vision tasks, including image classification, object detection and panoptic segmentation, we also empirically observe some frequent training crashes (i.e., model diverged with loss=nan) when we insert the BatchFormerV2 module into very early layers, especially in image classification task. In future, we will further explore how to smoothly use the BatchFormerV2 module in different layers of image classification models.

\section{Conclusion}

Attention mechanisms have attracted intensive interests from the communities of  natural language processing and computer vision. Previous approaches mainly investigate 1) self-attentions on either channel or spatial dimensions; 2) batch attentions for image-level sample relationship learning under data scarce settings. In this paper, we present a more general batch Transformer module, termed as BatchFormerV2, to explore sample relationships for dense representation learning. Specifically, the proposed BatchFormerV2 module can be easily integrated into existing vision Transformer architectures for sample relationship modeling within each training mini-batch from either pixel/patch- or image-levels. Meanwhile, we further propose a two-stream training pipeline for BatchFormerV2, where two streams share all other layers/blocks except the BatchFormerV2 modules. By doing this, BatchFormerV2 can thus be a plug-and-play module and easily integrated into different vision Transformers without introducing any extra inference cost.  Extensive experiments on a variety of visual recognition tasks, including image classification, objection detection, and panoptic segmentation, demonstrate the effectiveness of the proposed BatchFormerV2 module for robust representation learning.






%
%
\bibliographystyle{splncs04}
\bibliography{egbib}

\appendix

\section{Additional Experimental Details}
In this section, we provide more details about the image classification experiments in the main paper.
For ImageNet, we observe frequent training crashes (i.e., model diverged with nan loss) when we insert the BatchFormerV2 module into very early layers. Therefore, we insert the BatchFormerV2 module in the eighth layer. Specifically, for DeiT-B, we find that the model with the BatchFormerV2 module is more prone to collapse, possibly because the model is easier to overfit the data or batch patterns after we insert the BatchFormerV2 module into the models. Therefore, we use a larger weight decay (\ie, 0.5) for the BatchFormerV2 module for DeiT-B. Empirically, we find that a large weight decay is a simple yet effective solution to avoid the collapse during optimization. For CIFAR-100, we follow all default hyper-parameters in ~\cite{chen2021empirical}, except that we train all models from scratch. Specifically, the running script based on the released official code\footnote{\url{https://github.com/facebookresearch/moco-v3/tree/main/transfer}} of DeiT~\cite{touvron2021deit} is as follows,

\begin{figure}[!ht]
\centering
\begin{lstlisting}[language=Python]
python -u -m torch.distributed.launch --nproc_per_node=8 --use_env main.py \
    --batch-size 128 --output_dir [your output dir path] --epochs 100 --lr 3e-4 --weight-decay 0.1 \
    --no-pin-mem  --warmup-epochs 3 --data-set cifar100 --data-path [cifar-100 data path]  --no-repeated-aug \
    --reprob 0.0 --drop-path 0.1 --mixup 0.5 --cutmix 1 \
    --add_bt 1
\end{lstlisting}
\caption{Running script of BatchFormerV2 on CIFAR-100.}
\label{fig:command}
\end{figure}


\section{More Experimental Results}

In this section, we provide more experimental results on more visual recognition tasks, including long-tailed recognition, 3d hand reconstruction, and self-supervised learning, to demonstrate the good generalizability of the proposed BatchFormerV2.

\subsection{Long-Tailed Recognition}

In Table~\ref{tab:imagenet_lt}, we show the model performances with the BatchFormerV2 module on ImageNet-LT. Here, all experiments are based on DeiT-S~\cite{touvron2021deit} and we do not use any re-balance strategies. We find that the proposed BatchFormerV2 module can significantly improve the model performance comparing with the corresponding baseline.

\begin{table}[tp]
\small
\setlength\tabcolsep{3.5pt}
\centering
\begin{tabular}{@{}lccccc@{}}
\hline
Method & Insert Position & All & Many & Med & Few \\
\hline
Deit-S & - & 32.8 & 52.5 & 24.3 & 7.0  \\
+ BatchFormerV2 & 1-12 & {\bf 35.5} &  {\bf 55.4} & {\bf 27.2} & {\bf 8.6} \\
+ BatchFormerV2 & 8-12 & 34.7 &  54.7 & 26.3 & 7.2 \\
+ BatchFormerV2 & 4-12 & 35.5 &  55.3 & 26.8 & 8.4 \\
+ BatchFormerV2 (non-shared) & 1-12 & 35.2 & 55.3 & 26.7  & 8.3\\
\hline
\end{tabular}
\vspace{2mm}
\caption{Illustration of Deit-S on ImageNet-LT. By default, we share all the modules among different layers on this experiments. BatchFormerV2 (non-shared) indicates we do not share the modules among different layers. We observe sharing BatchFormerV2 on image classification achieves a bit better performance.}
\label{tab:imagenet_lt}
\end{table}

\subsection{3D Hand Reconstruction}

In addition to object detection and panoptic segmentation, we further provide results on another important pixel-level task, i.e., 3D Reconstruction. Specifically, we use the popular 3D hand reconstruction benchmark, \ie, FreiHAND~\cite{zimmermann2019freihand} and evaluate the proposed BatchFormerV2 module for 3D hand mesh reconstruction using a recent state-of-the-art method~\cite{lin2021-mesh-graphormer}, MeshGraphormer. Here, we report the performance on FreiHand dataset under single-scale inference for a quick evaluation.
As shown in Table~\ref{tab:3d_hand}, the proposed BatchFormerV2 module clearly improves the baseline by over 1.\% on both two metrics, PA-MPVPE and PA-MPJPE.

\begin{table}[!ht]
\small
\setlength\tabcolsep{2.5pt}
\centering
\begin{tabular}{@{}lcccc@{}}
\hline
Method & PA-MPVPE ↓ & PA-MPJPE ↓ & F@5 mm ↑ & F@15 mm ↑\\
\hline
Mesh Graphormer~\cite{lin2021-mesh-graphormer}* & 62.8 &  64.3 & 74.7 & 98.3 \\
+ BatchFormerV2 & {\bf 61.3} & {\bf 62.6} & {\bf 75.4} & {\bf 98.5}\\

\hline
\end{tabular}
\vspace{2mm}
\caption{BatchFormerV2 for 3D Hand Mesh Reconstruction. $*$ indicates we train the network with the released official code of~\cite{lin2021-mesh-graphormer}.}
\label{tab:3d_hand}
\end{table}

\begin{table}[!ht]
\small
\setlength{\tabcolsep}{5.5pt}
\centering
\begin{tabular}{@{}lccccccc@{}}
\toprule
Method & Epochs & ViT-Base & ViT-Large\\
\hline
MAE~\cite{he2021masked}* & 800 & 65.6 & 73.5 \\
+BatchFormerV2 & 800 & {\bf 66.1} & {\bf 73.9} \\
\bottomrule
\end{tabular}
\vspace{2mm}
\caption{MAE with BatchFormerV2. * indicates we use the released code to MAE for 800 epochs. We illustrate the result of Linear Probe.}
\label{tab:mae}
\end{table}

\subsection{Self-Supervised Learning}

Here, we also utilize a simple experiment to evaluate BatchFormerV2 on recent self-supervised learning framework, \ie, Masked Auto Encoder~\cite{he2021masked}(MAE). We insert BatchFormerV2 into all layers in the decoder in MAE~\cite{he2021masked}. We use the ViT-Base model to evaluate BatchFormerV2. Here, due to the computation limitation, we train the network 800 epochs with the released code of~\cite{he2021masked}, and verify the model via linear probe. All other hyper-parameters are following~\cite{he2021masked}.
Table~\ref{tab:mae} demonstrates BatchFormerV2 is also beneficial for MAE. Without bells and whistles, BatchFormerV2 improves the baseline by 0.5\%.

\section{Additional Ablation Studies}

\subsection{Without Two-stream Strategy}
We conduct experiments about the two-stream training strategy. As shown in  Table~\ref{tab:ab_two_stream}, the performance significantly drops if we use a single stream with BatchFormerV2, since the distribution between with and without BatchFormerV2 changes in each layer. Therefore, a single-stream network can not enable the inference without BatchFormerV2 modules.

\begin{table}[!ht]
\setlength{\tabcolsep}{5.5pt}
\centering
\begin{tabular}{@{}lccccccc@{}}
\toprule
Method & Backbone & $AP$ & $AP_{50}$ & $AP_{75}$ & $AP_S$ & $AP_M$ & $AP_L$ \\
\hline
w/ Two-stream Strategy & ResNet-50 & {\bf 45.5} & {\bf 64.3} & {\bf 49.8} & {\bf 28.3} & {\bf 48.6} & {\bf 59.4} \\
w/o Two-stream Strategy & ResNet-50 & 12.3 & 33.9 & 6.3 & 5.3 & 21.1 & 14.6 \\
\bottomrule
\end{tabular}
\vspace{2mm}
\caption{Ablation study on two-stream training strategy.}
\label{tab:ab_two_stream}
\end{table}

\subsection{Mini-batch Inference}

In our experiment, we remove BatchFormerV2 modules for inference, since we can not always assume a mini-batch of testing data. In Table~\ref{tab:mini_batch}, we also show the results of BatchFormerV2 with mini-batch inference. Here, we insert BatchFormerV2 module in the first layer, and use the optimized model to evaluation with mini-batch inference. We find that ``inference without BatchFormerV2" achieves similar performance comparing with ``inference with BatchFormerV2". Therefore, we think that the two-stream strategy enables the semantically invariant learning, and can remove BatchFormerV2 during inference.

\setlength{\tabcolsep}{5.5pt}
\begin{table}
\centering
\begin{tabular}{@{}lccccccc@{}}
\toprule
Method & Backbone & $AP$ & $AP_{50}$ & $AP_{75}$ & $AP_S$ & $AP_M$ & $AP_L$ \\
\hline
BatchFormerV2 & ResNet-50 &  45.6 & 64.5 & 49.8 & 28.3 & 48.8 & 59.7\\
+ Mini-batch Inference & ResNet-50 & 45.6 & 64.4 & 49.8 & 28.3 & 48.7 & 59.7 \\
\bottomrule
\end{tabular}
\vspace{2mm}
\caption{Ablation study on mini-batch inference with BatchFormerV2 modules.}
\label{tab:mini_batch}
\end{table}

\begin{table}
\centering
\setlength{\tabcolsep}{6pt}
\begin{tabular}{@{}lcccc@{}}
\toprule
Model & Params & Input & Top-1 & Top-5 \\
\hline

DeiT-S~\cite{touvron2021deit} & 22M & 224$^2$ & 81.8 & 94.1 \\
+ BatchFormerV2& 22M & 224$^2$ & {\bf 82.9} & {\bf 94.3} \\

DeiT-S~\cite{touvron2021deit} (w/o mixup) & 22M & 224$^2$ & 75.80 & 89.6 \\
+ BatchFormerV2 (w/o mixup) & 22M & 224$^2$ & {\bf 78.2} & {\bf 95.6} \\

DeiT-S~\cite{touvron2021deit} (w/o cutmix) & 22M & 224$^2$ & 79.8 & 92.5 \\
+ BatchFormerV2 (w/o cutmix) & 22M & 224$^2$ & {\bf 81.1} & {\bf 92.3} \\

\bottomrule
\end{tabular}
\vspace{2mm}
\caption{Illustration the effect of mixup~\cite{zhang2018mixup} and cutmix~\cite{yun2019cutmix} on BatchFormerV2. Experiments are conducted on Tiny-ImageNet. We follow the same experimental setups described in DeiT~\cite{touvron2021deit}. ``w/o mixup" indicates we remove both mixup and cutmix.}
\label{tab:img_tiny}
\end{table}

\subsection{Classification Without Mixup}

We notice that there are frequent training crashes when applying BatchFormerV2 with multiple layers on large datasets. We also evaluate BatchFormerV2 without mixup on Tiny-ImageNet. Except for BatchFormerV2 modules, all configurations follow ~\cite{touvron2021deit}. We run the experiments on four Nvidia V100 (16GB) GPUs. Table~\ref{tab:img_tiny} demonstrates that without using cutmix~\cite{yun2019cutmix} or mixup~\cite{zhang2018mixup}, BatchFormerV2 significantly improves the baseline with a larger margin.

\subsection{Shared BatchFormerV2 Modules}

We can also share BatchFormerV2 modules among different layers on image classification. The motivation behind of this setting is that we further encourage different layers to discover the same batch attention pattern. Here, we illustrates the effect of sharing modules among different layers on ImageNet-LT. As shown in Table~\ref{tab:imagenet_lt}, it achieves a bit better performance on small datasets if we share the modules among different layers. This is different from the observation on object detection, possibly because that sharing modules plays a role of regularization which benefits the learning on small datasets. Meanwhile, it is also challenging to optimize the DeiT model with BatchFormerV2 modules if we do not share the modules among different layers. In this paper, we mainly focus on a general BatchFormerV2 module which can be well generalized for different levels of tasks. We leave the further exploration of sharing strategy, and crash collapse on ImageNet when inserting BatchFormerV2 into multiple layers to future work.



\section{Visual Comparison}

\subsection{Visualization of Feature Representation}

\begin{figure}
    \centering
    \includegraphics[width=.89\textwidth]{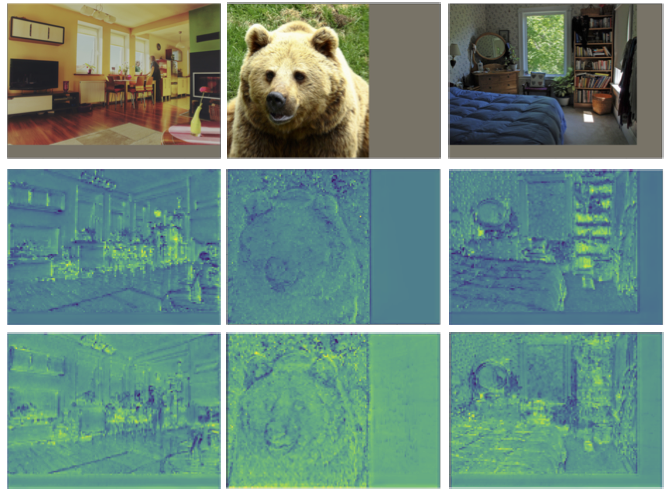}
    \caption{Visualization of the difference between the representations with BatchFormerV2 and without BatchFormerV2 during inference. Here, we choose the largest feature map and use the model that we trained with BatchFormerV2 which is inserted into the first Transformer Encoder layer. The first row is image, the second row is the feature without BatchFormerV2, and the last row indicates the feature with BatchFormerV2 (mini-batch inference).}
    \label{fig:bt_diff}
\end{figure}

Table~\ref{tab:mini_batch} shows BatchFormerV2 without mini-batch inference achieves similar performance to that with mini-batch inference. To further analyze this phenomenon, we visualize the feature maps between with BatchFormerV2 and without BatchFormerV2. As shown in Figure~\ref{fig:bt_diff}, we find that there are significant differences (\ie, different distribution) between the above-mentioned two feature maps during inference. We think the two feature maps represent similar semantics though the distribution is diverse, \ie, representing similar semantics for the the same prediction modules.

\subsection{Visualization of Panoptic Segmentation}
\begin{figure}
    \centering
    \includegraphics[width=.95\textwidth]{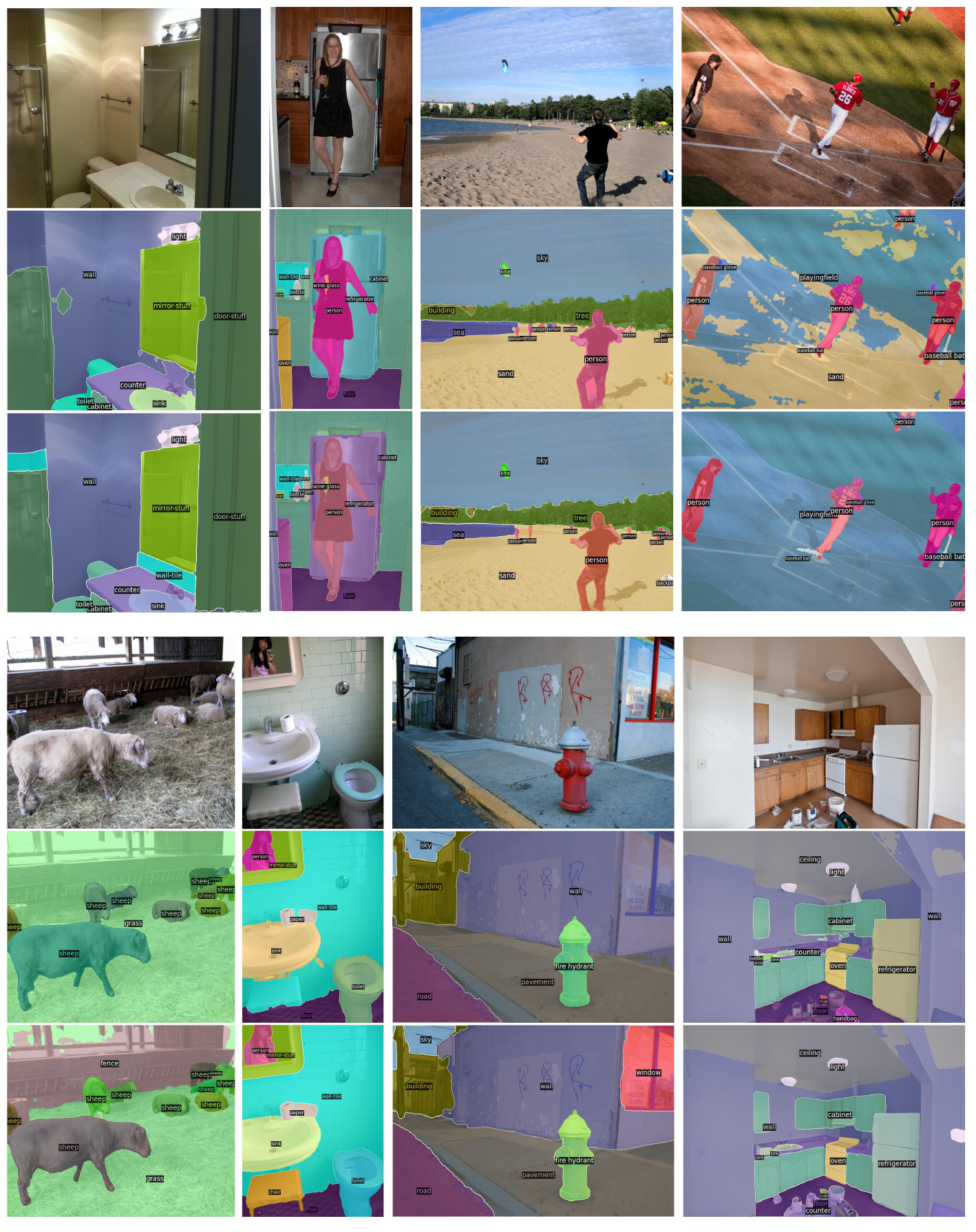}
    \caption{Visualization (1) of additional panoptic segmentation examples. The first row is original image, the second row is DETR and the thrid row is DETR with BatchFormerV2.}
    \label{fig:bt2_panop_seg1}
\vspace{-2mm}
\end{figure}

We further provide more panoptic segmentation examples in Figure~\ref{fig:bt2_panop_seg1}. We find that BatchFormerV2 usually helps object segmentation and improves the segmentation boundaries of the stuffs.


\subsection{Visualization of Attention}

\begin{figure}
    \centering
    \includegraphics[width=.85\textwidth]{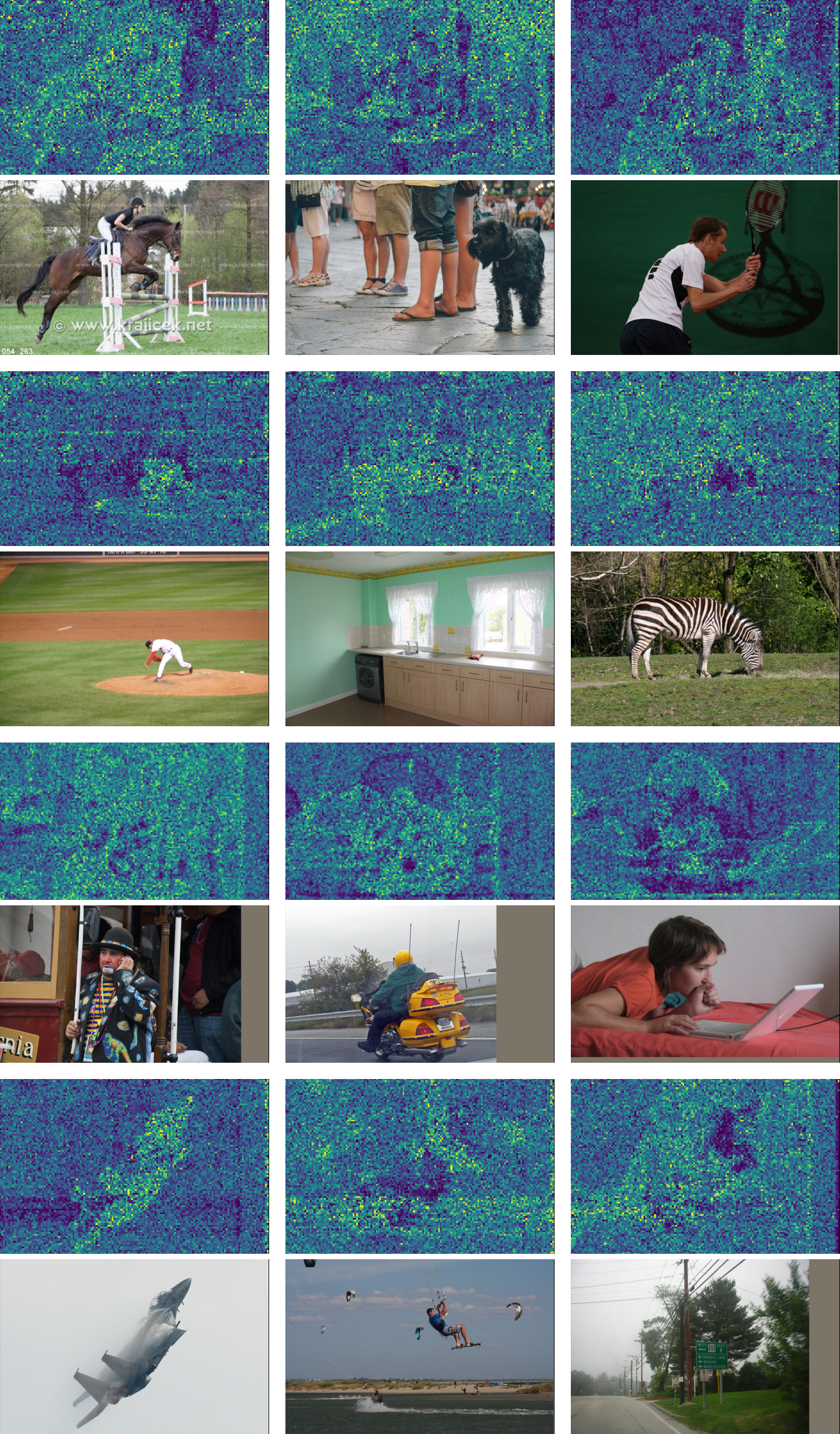}
    \caption{Visualization of self-attention in the same mini-batch. Each row represents a mini-batch during inference. The model and settings are the same as those in Figure 5 in main paper}
    \label{fig:bt_batch_atten}
\end{figure}

Visualization of the multi-head self-attention provides rich semantic interpretations. Here we provide more observations from the visualization of attentions in Figure~\ref{fig:bt_batch_atten}. First, we observe that the images with objects usually have higher attentions to other images, \ie, the objects are usually highlighted as illustrated in Figure~\ref{fig:bt_batch_atten}. Second, and more importantly, \textit{the attention of background in current image is suppressed if the corresponding positions in other images have objects}. For example, the region (grass) under the zebra in row 2 in Figure~\ref{fig:bt_batch_atten} is suppressed because there is a person in the first image. There is a region suppressed like a person in ``row 4, column 3" in Figure~\ref{fig:bt_batch_atten} because there is a person in second column. However, if the region has objects, the region will not be suppressed. For example, the airplane is highlighted in ``row 4, column 1" though the corresponding region is object in ``row 4, column 2".

\end{document}